\documentclass[11pt,a4paper]{article}
\usepackage[hyperref]{acl2018}
\usepackage{times}
\usepackage{latexsym}
\usepackage{booktabs}
\usepackage{multirow}

\usepackage{tikz}\usepackage{tikz-qtree}

\usepackage{color}
\definecolor{strawberry}{RGB}{219, 73, 76}
\definecolor{goldenrod}{RGB}{235, 164, 20}
\definecolor{leafgreen}{RGB}{134, 170, 109}
\definecolor{skyblue}{RGB}{70, 181, 211}

\usepackage{url}

\aclfinalcopy 
\def\aclpaperid{937} 

\title{Deep RNNs Encode Soft Hierarchical Syntax}

\author{Terra Blevins, Omer Levy, and Luke Zettlemoyer \\
        Paul G. Allen School of Computer Science \& Engineering\\ University of Washington\\ Seattle, WA \\ {\tt \{blvns, omerlevy, lsz\}@cs.washington.edu}}

\date{}

\begin{document}
\maketitle

\begin{abstract}
We present a set of experiments to demonstrate that deep recurrent neural networks (RNNs) learn internal representations that capture soft hierarchical notions of syntax from highly varied supervision. We consider four syntax tasks at different depths of the parse tree; for each word, we predict its part of speech as well as the first (parent), second (grandparent) and third level (great-grandparent) constituent labels that appear above it. These predictions are made from representations produced at different depths in networks that are pretrained with one of four objectives: dependency parsing, semantic role labeling, machine translation, or language modeling. In every case, we find a correspondence between network depth and syntactic depth, suggesting that a soft syntactic hierarchy emerges. This effect is robust across all conditions, indicating that the models encode significant amounts of syntax even in the absence of an explicit syntactic training supervision.
\end{abstract}

\section{Introduction}
Deep recurrent neural networks (RNNs) have effectively replaced explicit syntactic features (e.g. parts of speech, dependencies) in state-of-the-art NLP models \cite{he2017,lee2017,opennmt}. However, previous work has shown that syntactic information (in the form of either input features or supervision) is useful for a wide variety of NLP tasks \cite{punyakanok2005,chiang2009}, even in the neural setting \cite{aharoni2017,chen2017}. In this paper, we show that the internal representations of RNNs trained on a variety of NLP tasks encode these syntactic features without explicit supervision. 

We consider a set of feature prediction tasks drawn from different depths of syntactic parse trees; given a word-level representation, we attempt to predict the POS tag and the parent, grandparent, and great-grandparent constituent labels of that word. We evaluate how well a simple feed-forward classifier can detect these syntax features from the word representations produced by the RNN layers from deep NLP models trained on the tasks of dependency parsing, semantic role labeling, machine translation, and language modeling. We also evaluate whether a similar classifier can predict if a dependency arc exists between two words in a sentence, given their representations.

We find that, across all four types of supervision, the representations learned by these models encode syntax beyond the explicit information they encounter during training; this is seen in both the word-level tasks and the dependency arc prediction task. Furthermore, we also observe that features associated with different levels of syntax tree correlate with word representations produced by RNNs at different depths. Largely speaking, we see that deeper layers in each model capture notions of syntax that are higher-level and more abstract, in the sense that higher-level constituents cover a larger span of the underlying sentence.

These findings suggest that models trained on NLP tasks are able to induce syntax even when direct syntactic supervision is unavailable. Furthermore, the models are able to differentiate this induced syntax into a soft hierarchy across different layers of the model, perhaps shedding some light on why \textit{deep} RNNs are so useful for NLP.

\section{Methodology}

\begin{table*}[t]
\small
\centering
\begin{tabular}{lllcc}
\toprule
\multirow{2}{*}{\textbf{Training Signal}} & \multirow{2}{*}{\textbf{Dataset}} & \multirow{2}{*}{\textbf{RNN Layers}} & \textbf{Input} & \textbf{Hidden} \\
 & & & \textbf{Dims} & \textbf{Dims} \\
\midrule
\multirow{2}{*}{Dependency Parsing} & English Universal &  4 parallel bidirectional & \multirow{2}{*}{200} & \multirow{2}{*}{400} \\
 &  Dependencies & LSTMs & & \\
\midrule
\multirow{2}{*}{Semantic Role Labeling} & \multirow{2}{*}{CoNLL-2012} & 8 alternating-direction & \multirow{2}{*}{100} & \multirow{2}{*}{300} \\
 & & LSTMs with highways & & \\
\midrule
\multirow{2}{*}{Machine Translation} & WMT-2014 & 4 parallel bidirectional & \multirow{2}{*}{500} & \multirow{2}{*}{500} \\
 & English-German & LSTMs & & \\
\midrule
\multirow{2}{*}{Language Modeling} & \multirow{2}{*}{CoNLL-2012} & 2 sets of 4 unidirectional & \multirow{2}{*}{1000} & \multirow{2}{*}{1000} \\
 & & LSTMs with highways & & \\
\bottomrule	
\end{tabular}
\caption{The training data, recurrent architecture, and hyperparameters of each model.}
\label{tab:dimensions}
\end{table*}

\begin{figure}
	\begin{tikzpicture}[scale=0.85,
	level distance=30pt,
	frontier/.style={distance from root=150pt},
	every leaf node/.append style={text depth=0pt}]
  \Tree [.S [.NP [.JJ Other ]
                 [.NN stock ]
                 [.NNS indexes ] ]
            [.ADVP [.RB also ] ]
            [.\textcolor{strawberry}{\textbf{VP}} [.VBD fell ]
                   [.\textcolor{goldenrod}{\textbf{PP}} [.IN on ] 
                        [.\textcolor{skyblue}{\textbf{NP}} [.\textcolor{leafgreen}{\textbf{NNP}} \textit{Monday} ] ] ] ] ]
	\end{tikzpicture}
    \vspace{-5pt}
	\caption{Constituency tree with labels for the word ``Monday'' for the POS (green), parent constituent (blue), grandparent constituent (orange), and great-grandparent constituent (red) tasks.}
	\label{constituency-tree}
\end{figure}
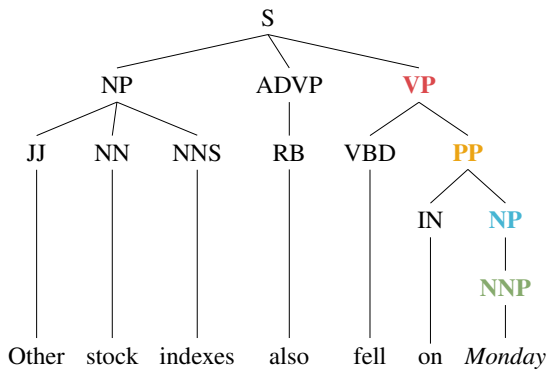

Given a model that uses multi-layered RNNs, we collect the vector representation $\mathbf{x}^l_i$ of each word $i$ at each hidden layer $l$. To determine what syntactic information is stored in each word vector, we try to predict a series of constituency-based properties from the vector alone. Specifically, we predict the word's part of speech (POS), as well as the first (parent), second (grand-parent), and third level (great-grandparent) constituent labels of the given word. Figure \ref{constituency-tree} shows how these labels correspond to an example constituency tree.

Our methodology follows \citet{shi2016}, who run syntactic feature prediction experiments over a number of different shallow machine translation models, and Belinkov et al. \citeyear{belinkov2017,belinkov2017b}, who use a similar process to study the morphological, part-of-speech, and semantic features learned by deeper machine translation encoders. We extend upon prior work by considering training signals for models other than machine translation, and by applying more stratified word-level syntactic tasks.

\subsection{Experiment Setup}
We predict each syntactic property with a simple feed-forward network with a single 300-dimensional hidden layer activated by a ReLU:
\begin{equation}
\mathbf{y}^l_i = \textnormal{SoftMax} (W_2 \textnormal{ReLU} (W_1 \mathbf{x}^l_i))
\end{equation}
where $i$ is the word index and $l$ is the layer index within a model. To ensure that the classifiers are not trained on the same data as the RNNs, we train the classifier for each layer $l$ separately using the development set of CoNLL-2012  and evaluate on the test set \cite{pradhan2013}.

In addition, we compare performance with  word-level baselines. We report the per-word majority class baseline; at the POS level, for example, ``cat'' will be classified as a noun and ``walks'' as a verb. This baseline outperforms the pre-trained GloVe \cite{pennington2014} embeddings on every task. We also consider a contextual baseline, in which we concatenate each word's embedding with the average of its context's embeddings; however, this baseline also performed worse that the reported one.

\begin{figure*}[t]
  \minipage{0.5\textwidth}
    \includegraphics[width=\linewidth]{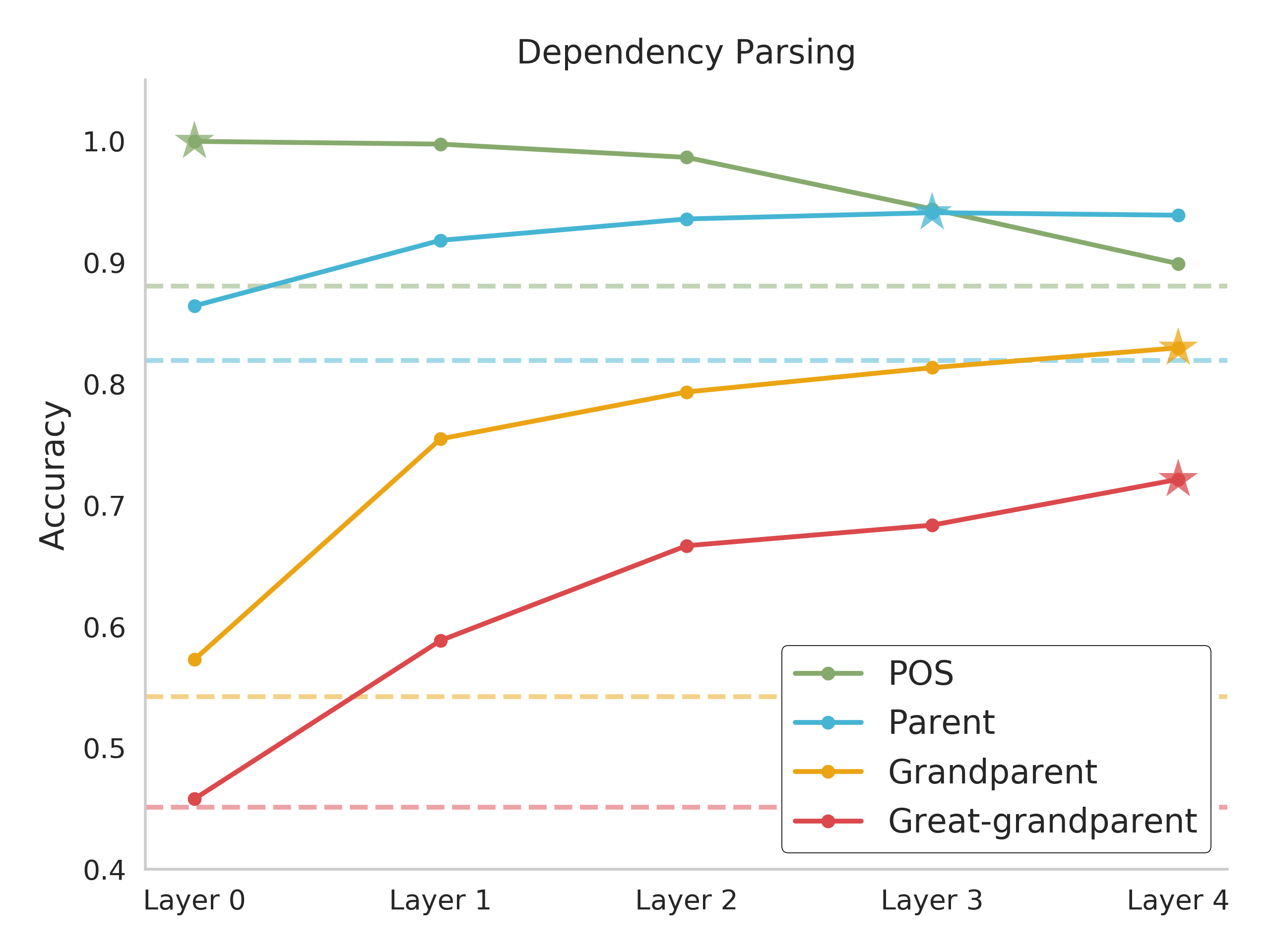}
  \endminipage\hfill
  \minipage{0.5\textwidth}
    \includegraphics[width=\linewidth]{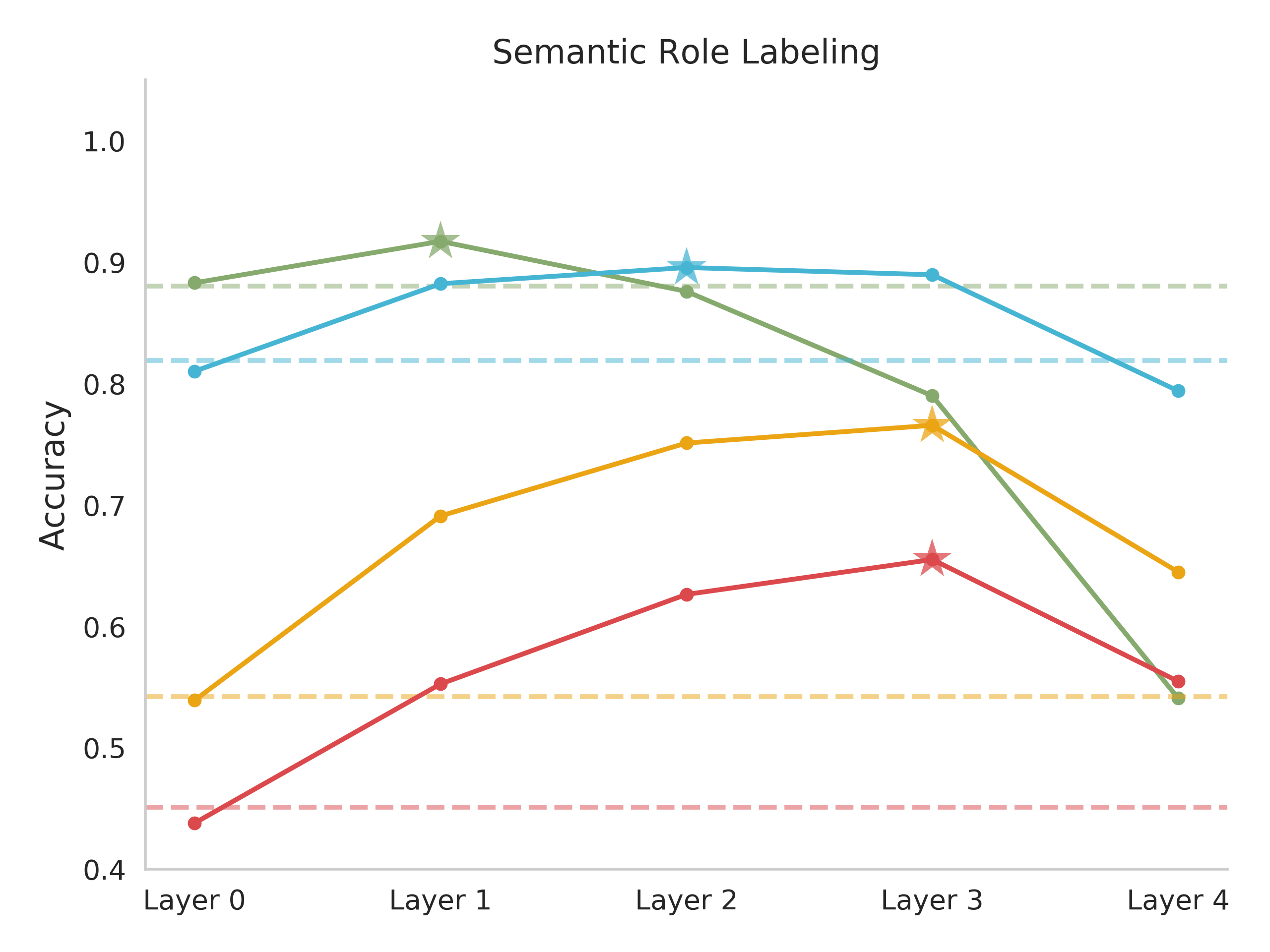}
  \endminipage\hfill
  \minipage{0.5\textwidth}
    \includegraphics[width=\linewidth]{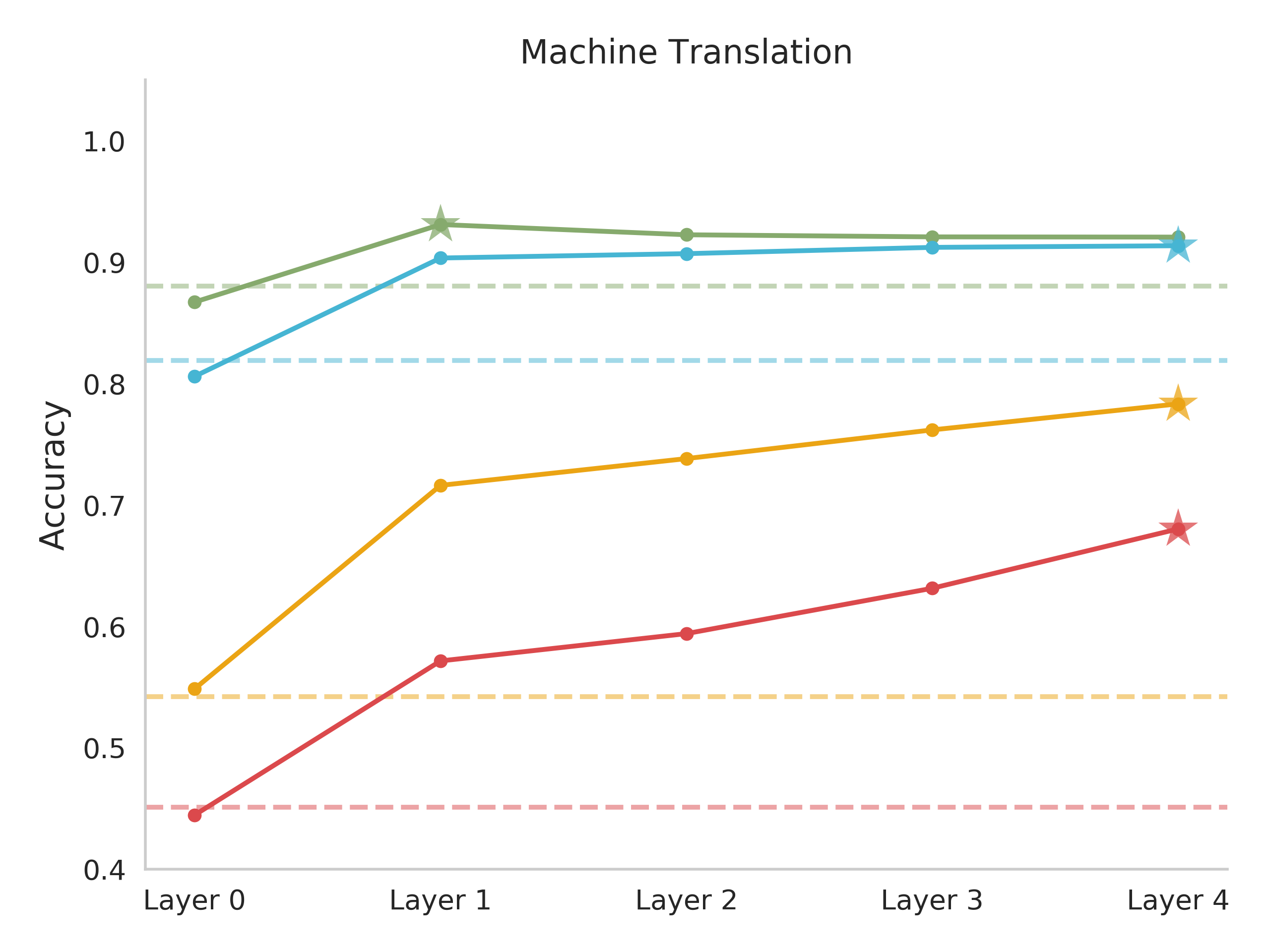}
  \endminipage\hfill
  \minipage{0.5\textwidth}
  	\includegraphics[width=\linewidth]{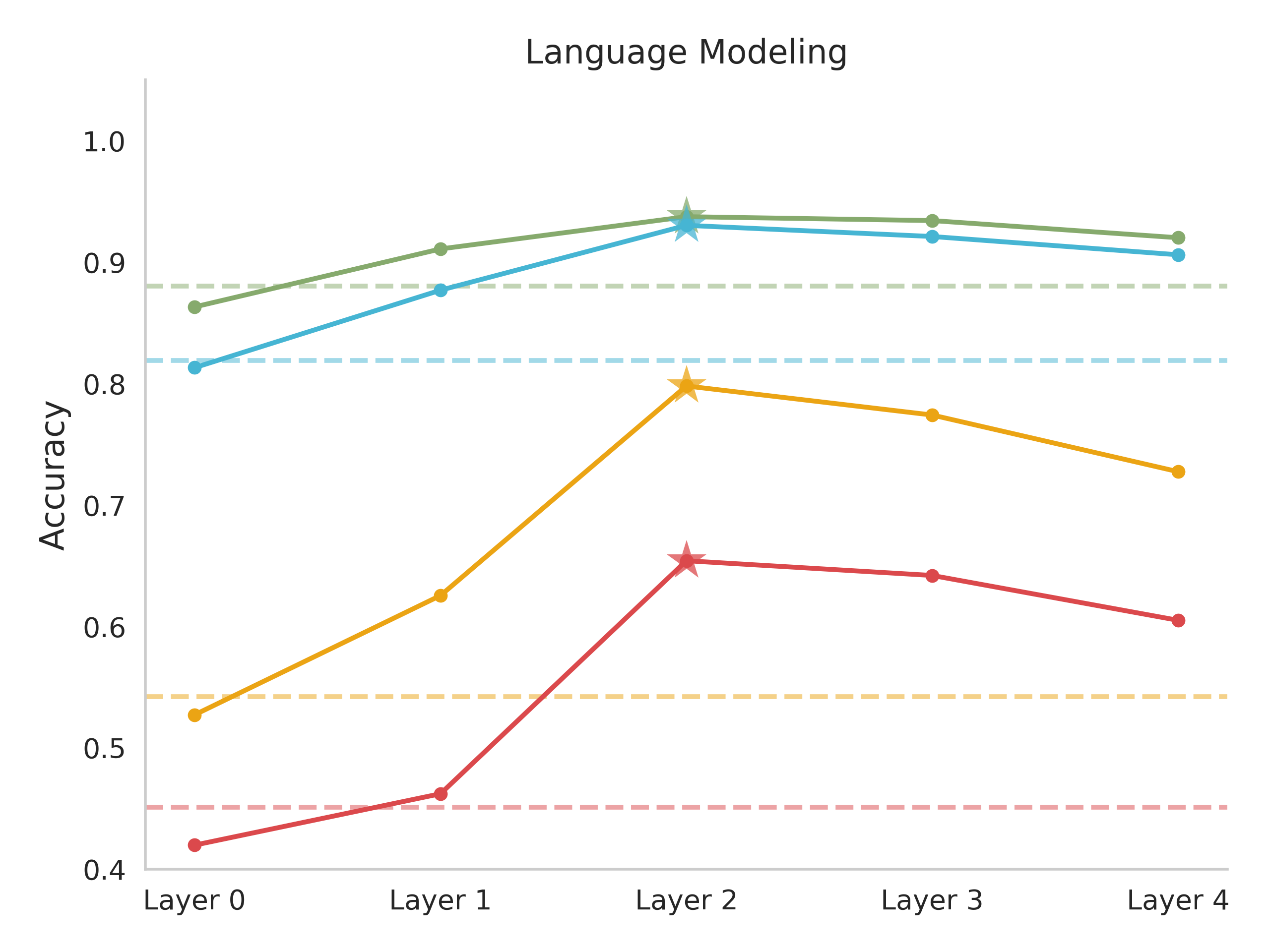}
  \endminipage
  \caption{Results of syntax experiments. The best performing layer for each experiment is annotated with a star, and the per-word majority baseline for each task is shown with a dashed line.}
  \label{results}
\end{figure*}

\subsection{Analyzed Models}

We consider four different forms of supervision. Table~\ref{tab:dimensions} summarizes the differences in data, architecture, and hyperparameters.\footnote{While we control for some variables, we  mainly rely on existing architectures and hyperparameters that were tuned for the original tasks, limiting the cross-model comparability of absolute performance levels on our syntactic evaluations.}

\paragraph{Dependency Parsing}
We train a four-layer version of the Stanford dependency parser \cite{dozat17} on the Universal Dependencies English Web Treebank \cite{silveira14}. We ran the parser with 4 bidirectional LSTM layers (the default is 3), yielding a UAS of 91.5 and a weighted LAS of 82.18, consistent with the state of the art on CoNLL 2017.
Since the parser receives syntactic features as input (POS) and is trained on an explicit syntactic signal, we expect its intermediate representations to contain a high amount of syntactic information.

\paragraph{Semantic Role Labeling}
We use the pre-trained DeepSRL model from \cite{he2017}, which was trained on the training data from the CoNLL-2012 dataset. This model is an alternating bidirectional LSTM, where the model consists of eight total layers that alternate between a forward layer and backward layer. We concatenate the representations from each pair of directional layers in the model for consistency with other models.

\paragraph{Machine Translation}
We train a machine translation model using OpenNMT \cite{opennmt} on the WMT-14 English-German dataset. The encoder (which we examine in our experiments) is a 4-layer bidirectional LSTM; we use the defaults for every other setting. The model achieves a BLEU score of 21.37, which is in the ballpark of other vanilla encoder-decoder attention models on this benchmark \cite{bahdanau2014neural}.

\paragraph{Language Modeling}
We train two separate language models on CoNLL-2012's training set, one going forward and another backward. Each model is a 4-layer LSTM with highway connections, variational dropout, and tied input-output embeddings.
After training, we concatenate the forward and backward representations for each layer.\footnote{The model achieved perplexities of 50.56 (forward) and 51.24 (backward) on CoNLL-2012's test set. Since we are not familiar with other perplexity results on this data, we note that retraining the architecture on Penn TreeBank achieved 64.39 perplexity, which is comparable to other high-performing language models.} 

\section{Constituency Label Prediction}

Figure~\ref{results} shows our results (see supplementary material for numerical results).
We make several observations:

\paragraph{RNNs can induce syntax.}
Overall, each model outperforms the baseline and its respective input embeddings on every syntax task, indicating that their internal representations encode some notions of syntax. The only exception to this observation is POS prediction with dependency parsing representations; in this case the parser is provided gold POS tags as input, and cannot improve upon them. This result confirms the findings of \citet{shi2016} and \citet{belinkov2017b}, who demonstrate that neural machine translation encoders learn syntax, and shows that RNNs trained on other NLP tasks also induce syntax.

\paragraph{Deeper layers reflect higher-level syntax.}
In 11 out of 16 cases, performance improves up to a certain layer and then declines, suggesting that the deeper layers encode less syntactic information that earlier ones in these cases. 
Strikingly, the higher-level a syntactic task is, the deeper in the network the peak performance occurs; for example, in SRL we see that the parent constituent task peaks one layer after POS, and the grand-parent and great-grandparent tasks peak on the layer after that.
One possible explanation is that each layer leverages the shallower syntactic information learned in the previous layer in order to construct a more abstract syntactic representation.
In SRL and language modeling, it seems as though the syntactic information is then replaced by task-specific information (semantic roles, word probabilities), perhaps making it redundant.

This observation may also explain a modeling decision in ELMo \cite{peters2018}, where injecting the contextualized word representations from a pre-trained language model was shown to boost performance on a wide variety of NLP tasks. ELMo represents each word using a task-specific weighted sum of the language model's hidden layers, i.e. rather than use only the top layer, it selects which of the language model's internal layers contain the most relevant information for the task at hand. Our results confirm that, in general, different types of information manifest at different layers, suggesting that post-hoc layer selection can be beneficial.

\begin{figure}
  \minipage{0.5\textwidth}
    \includegraphics[width=\linewidth]{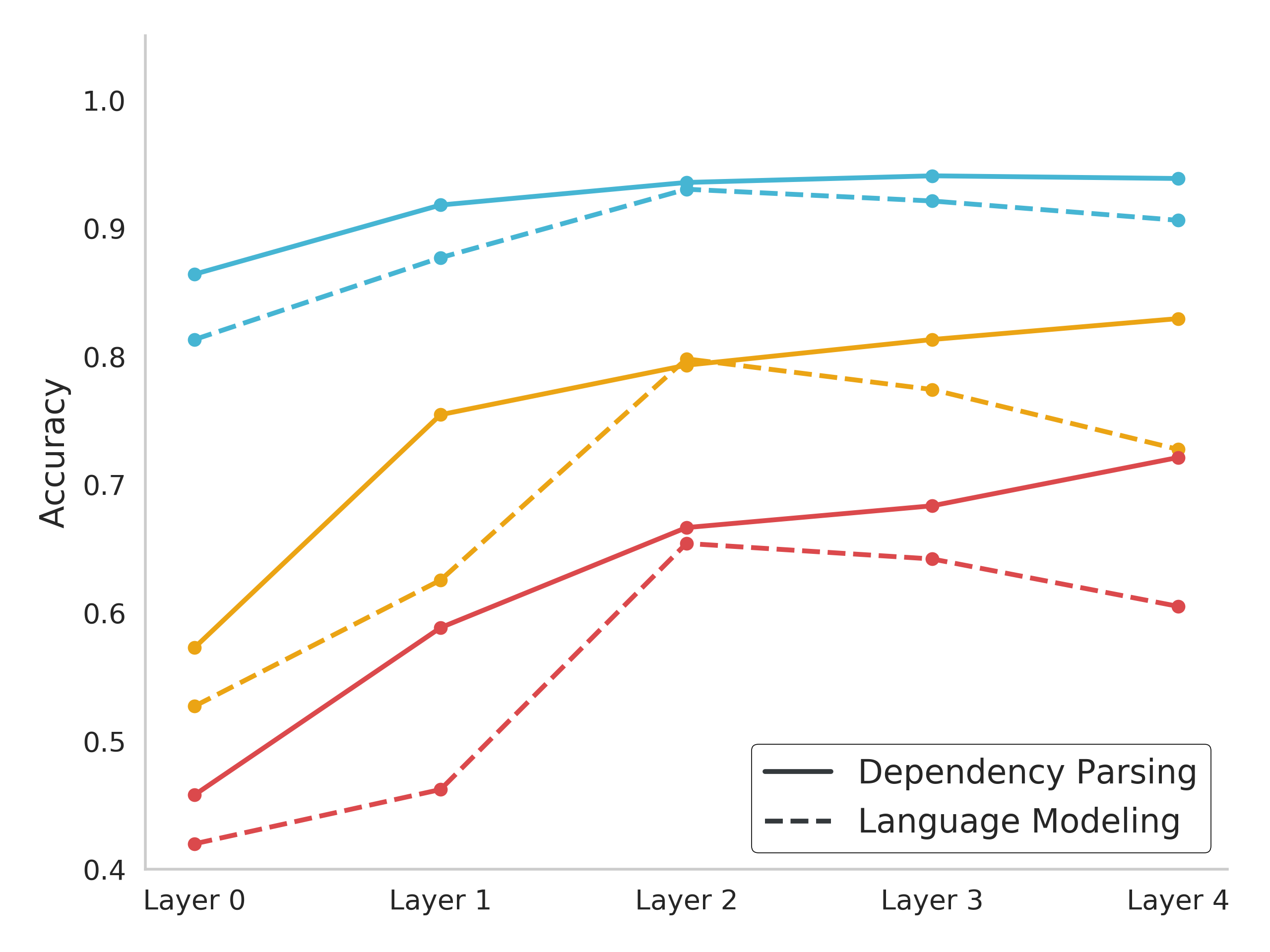}
  \endminipage
  \caption{Comparison between the LM and dependency parser on the parent (blue), grandparent (yellow), and great-grandparent (red) constituent prediction tasks.}
  \label{comparison}
\end{figure}

\paragraph{Language models learn some syntax.}
We compare the performance of language model representations to those learned with dependency parsing supervision, in order to gauge the amount of syntax induced. While this comparison is not ideal (the models were trained with slightly different architectures and hyperparameters), it does provide evidence that the language model's representations encode some amount of syntax implicitly. Specifically, we observe in Figure \ref{comparison} that the language model and dependency parser perform nearly identically on the three constituent prediction tasks in the second layer of their respective networks. 
In deeper layers the parser continues to improve, while the language model peaks at layer 2 and drops off afterwards.

These results may be surprising given the findings of \citet{linzen2016}, which found that RNNs trained on language modeling perform below baseline levels on the task of subject-verb agreement. However, the more recent investigation by \citet{gulordava2018} are in line with our results. They find that language models trained on a number of different languages assign higher probabilities to valid long-distance dependencies than to incorrect ones. Therefore, LMs seem able to induce syntactic information despite being provided with no linguistic annotation.

\begin{table*}
\small
\centering
  \begin{tabular}{ccccccc}
  \toprule
  \textbf{Source Model} & \textbf{GloVe} & \textbf{L0} & \textbf{L1} & \textbf{L2} & \textbf{L3} & \textbf{L4} \\
  \midrule
  DP & 0.50 & 0.68 & 0.77 & 0.81 & 0.88 & \textbf{0.95} \\
  SRL & 0.50 & 0.58 & 0.69 & 0.76 & \textbf{0.79} & 0.74 \\
  MT & 0.50 & 0.61 & \textbf{0.73} & 0.63 & 0.63 & 0.63 \\
  LM & 0.50 & 0.62 & 0.74 & 0.78 & \textbf{0.80} & 0.73 \\
  \bottomrule
  \end{tabular}
\caption{Results of the dependency arc prediction task. L0--L4 denote the different layers of the model. DP refers to the RNN trained with dependency parsing supervision.}
\label{tab:arc-pred}
\end{table*}

\section{Dependency Arc Prediction}
We run an additional experiment that seeks to clarify if the representations learned by deep NLP models capture information about syntactic structure. Using the internal representations from a deep RNN, we train a classifier to predict whether two words share an dependency arc (have a parent-child relationship) in the in the dependency parse tree over a sentence. We find that, similarly to the previous set of tasks, deep RNNs trained on various linguistic signals encode notions of the syntactic relationships between words in a sentence.

\paragraph{Setup} We use the same pretrained deep RNNs and feed-forward prediction network paradigm. However, we change the input from the previous experiments, as this task is not at the word-level, but rather concerns the relationship between two words; therefore, given a word pair $w_c$, $w_p$ for which we have a dependency arc label, we input $[w_c;w_p;w_c\circ w_p]$ into the classifier.

We use the Universal Dependencies dataset for this task, such that we train each classifier on the development set of this dataset and evaluate on the test set. We set up the task by generating two pairs of examples for each word in the UD dataset: a positive pair that consists of the word and its parent in the dependency tree, and a negative pair that matches the word with another randomly chosen word from the sentence.

\paragraph{Results} The results for this prediction task are given in Table \ref{tab:arc-pred}. We see the best performance from the dependency parser, finding that the performance for the dependency parser's representations continue to improve in the deepest layers, with a maximum performance of approximately 95\% on the last layer. This result is unsurprising, as this closely related to the task on which the model was explicitly trained. In the three other models, we find peaks that occur 12 to 20 accuracy points above the input layer's performance. These results support the findings from the constituency label prediction task and show that these findings hold up across syntactic formalisms. 

Similarly to the word-level tasks, we see the best performance from deeper layers in the models, with both SRL and LM performance peaking on the third layer. For the LM, we find that the best performing layer outperforms the initial layer by 18\%. This is consistent with our finding in the previous set of experiments, that RNNs encode significant amounts of syntax information even when trained on linguistic tasks without any explicit annotations. 

\section{Conclusions}

In this paper, we run a series of prediction tasks on the internal representations of deep NLP models, and find these RNNs are able to induce syntax without explicit linguistic supervision. We also observe that the representations taken from deeper layers of the RNNs perform better on higher-level syntax tasks than those from shallower layers, suggesting that these recurrent models induce a soft hierarchy over the encoded syntax. These results provide some insight as to why deep RNNs are able to model NLP tasks without annotated linguistic features. Further characterizing the exact aspects of syntax which these models can capture (and perhaps more importantly, those they cannot) is an interesting area for future work.

\section*{Acknowledgments}

The research was supported in part by the ARO (W911NF-16-1-0121) and the NSF (IIS-1252835, IIS-1562364). We also thank Yonatan Bisk, Yoav Goldberg, and the UW NLP group for helpful conversations and comments on the work.

\bibliography{acl2018}

\begin{thebibliography}{}
\expandafter\ifx\csname natexlab\endcsname\relax\def\natexlab#1{#1}\fi

\bibitem[{Aharoni and Goldberg(2017)}]{aharoni2017}
Roee Aharoni and Yoav Goldberg. 2017.
\newblock \href{https://doi.org/10.18653/v1/P17-2021}{Towards string-to-tree
  neural machine translation}.
\newblock In {\em Proceedings of the 55th Annual Meeting of the Association for
  Computational Linguistics (Volume 2: Short Papers)\/}. Association for
  Computational Linguistics, pages 132--140.
\newblock
  \href{https://doi.org/10.18653/v1/P17-2021}{https://doi.org/10.18653/v1/P17-2021}.

\bibitem[{Bahdanau et~al.(2015)Bahdanau, Cho, and Bengio}]{bahdanau2014neural}
Dzmitry Bahdanau, Kyunghyun Cho, and Yoshua Bengio. 2015.
\newblock \href{https://arxiv.org/abs/1409.0473}{Neural machine translation by
  jointly learning to align and translate}.
\newblock In {\em Proceedings of the Third International Conference on Learning
  Representations (ICLR)\/}.
\newblock
  \href{https://arxiv.org/abs/1409.0473}{https://arxiv.org/abs/1409.0473}.

\bibitem[{Belinkov et~al.(2017{\natexlab{a}})Belinkov, Durrani, Dalvi, Sajjad,
  and Glass}]{belinkov2017}
Yonatan Belinkov, Nadir Durrani, Fahim Dalvi, Hassan Sajjad, and James Glass.
  2017{\natexlab{a}}.
\newblock \href{https://doi.org/10.18653/v1/P17-1080}{What do neural machine
  translation models learn about morphology?}
\newblock In {\em Proceedings of the 55th Annual Meeting of the Association for
  Computational Linguistics (Volume 1: Long Papers)\/}. Association for
  Computational Linguistics, pages 861--872.
\newblock
  \href{https://doi.org/10.18653/v1/P17-1080}{https://doi.org/10.18653/v1/P17-1080}.

\bibitem[{Belinkov et~al.(2017{\natexlab{b}})Belinkov, M{\`a}rquez, Sajjad,
  Durrani, Dalvi, and Glass}]{belinkov2017b}
Yonatan Belinkov, Llu{\'i}s M{\`a}rquez, Hassan Sajjad, Nadir Durrani, Fahim
  Dalvi, and James Glass. 2017{\natexlab{b}}.
\newblock \href{http://aclweb.org/anthology/I17-1001}{Evaluating layers of
  representation in neural machine translation on part-of-speech and semantic
  tagging tasks}.
\newblock In {\em Proceedings of the Eighth International Joint Conference on
  Natural Language Processing (Volume 1: Long Papers)\/}. Asian Federation of
  Natural Language Processing, pages 1--10.
\newblock
  \href{http://aclweb.org/anthology/I17-1001}{http://aclweb.org/anthology/I17-1001}.

\bibitem[{Chen et~al.(2017)Chen, Fisch, Weston, and Bordes}]{chen2017}
Danqi Chen, Adam Fisch, Jason Weston, and Antoine Bordes. 2017.
\newblock \href{http://aclweb.org/anthology/P17-1171}{Reading wikipedia to
  answer open-domain questions}.
\newblock In {\em Proceedings of the 55th Annual Meeting of the Association for
  Computational Linguistics (Volume 1: Long Papers)\/}. Association for
  Computational Linguistics, Vancouver, Canada, pages 1870--1879.
\newblock
  \href{http://aclweb.org/anthology/P17-1171}{http://aclweb.org/anthology/P17-1171}.

\bibitem[{Chiang et~al.(2009)Chiang, Knight, and Wang}]{chiang2009}
David Chiang, Kevin Knight, and Wei Wang. 2009.
\newblock \href{http://www.aclweb.org/anthology/N09-1025}{11,001 new features
  for statistical machine translation}.
\newblock In {\em Proceedings of Human Language Technologies: The 2009 Annual
  Conference of the North American Chapter of the Association for Computational
  Linguistics\/}. Association for Computational Linguistics, pages 218--226.
\newblock
  \href{http://www.aclweb.org/anthology/N09-1025}{http://www.aclweb.org/anthology/N09-1025}.

\bibitem[{Dozat and Manning(2017)}]{dozat17}
Timothy Dozat and Christopher~D. Manning. 2017.
\newblock \href{http://arxiv.org/abs/1611.01734}{Deep biaffine attention for
  neural dependency parsing}.
\newblock In {\em Proceedings of the Fifth International Conference on Learning
  Representations (ICLR)\/}.
\newblock
  \href{http://arxiv.org/abs/1611.01734}{http://arxiv.org/abs/1611.01734}.

\bibitem[{Gulordava et~al.(2018)Gulordava, Bojanowski, Grave, Linzen, and
  Baroni}]{gulordava2018}
Kristina Gulordava, Piotr Bojanowski, Edouard Grave, Tal Linzen, and Marco
  Baroni. 2018.
\newblock \href{https://arxiv.org/pdf/1803.11138.pdf}{Colorless green recurrent
  networks dream hierarchically}.
\newblock In {\em Proceedings of NAACL-HLT\/}.
\newblock
  \href{https://arxiv.org/pdf/1803.11138.pdf}{https://arxiv.org/pdf/1803.11138.pdf}.

\bibitem[{He et~al.(2017)He, Lee, Lewis, and Zettlemoyer}]{he2017}
Luheng He, Kenton Lee, Mike Lewis, and Luke Zettlemoyer. 2017.
\newblock \href{https://doi.org/10.18653/v1/P17-1044}{Deep semantic role
  labeling: What works and what's next}.
\newblock In {\em Proceedings of the 55th Annual Meeting of the Association for
  Computational Linguistics (Volume 1: Long Papers)\/}. Association for
  Computational Linguistics, pages 473--483.
\newblock
  \href{https://doi.org/10.18653/v1/P17-1044}{https://doi.org/10.18653/v1/P17-1044}.

\bibitem[{Klein et~al.(2017)Klein, Kim, Deng, Senellart, and Rush}]{opennmt}
Guillaume Klein, Yoon Kim, Yuntian Deng, Jean Senellart, and Alexander Rush.
  2017.
\newblock \href{http://www.aclweb.org/anthology/P17-4012}{Opennmt: Open-source
  toolkit for neural machine translation}.
\newblock In {\em Proceedings of ACL 2017, System Demonstrations\/}.
  Association for Computational Linguistics, pages 67--72.
\newblock
  \href{http://www.aclweb.org/anthology/P17-4012}{http://www.aclweb.org/anthology/P17-4012}.

\bibitem[{Lee et~al.(2017)Lee, He, Lewis, and Zettlemoyer}]{lee2017}
Kenton Lee, Luheng He, Mike Lewis, and Luke Zettlemoyer. 2017.
\newblock \href{http://aclweb.org/anthology/D17-1018}{End-to-end neural
  coreference resolution}.
\newblock In {\em Proceedings of the 2017 Conference on Empirical Methods in
  Natural Language Processing\/}. Association for Computational Linguistics,
  pages 188--197.
\newblock
  \href{http://aclweb.org/anthology/D17-1018}{http://aclweb.org/anthology/D17-1018}.

\bibitem[{Linzen et~al.(2016)Linzen, Dupoux, and Goldberg}]{linzen2016}
Tal Linzen, Emmanuel Dupoux, and Yoav Goldberg. 2016.
\newblock \href{http://www.aclweb.org/anthology/Q16-1037}{Assessing the ability
  of {LSTMs} to learn syntax-sensitive dependencies}.
\newblock {\em Transactions of the Association of Computational Linguistics\/}
  4:521--535.
\newblock
  \href{http://www.aclweb.org/anthology/Q16-1037}{http://www.aclweb.org/anthology/Q16-1037}.

\bibitem[{Pennington et~al.(2014)Pennington, Socher, and
  Manning}]{pennington2014}
Jeffrey Pennington, Richard Socher, and Christopher Manning. 2014.
\newblock \href{https://doi.org/10.3115/v1/D14-1162}{Glove: Global vectors for
  word representation}.
\newblock In {\em Proceedings of the 2014 Conference on Empirical Methods in
  Natural Language Processing (EMNLP)\/}. Association for Computational
  Linguistics, pages 1532--1543.
\newblock
  \href{https://doi.org/10.3115/v1/D14-1162}{https://doi.org/10.3115/v1/D14-1162}.

\bibitem[{Peters et~al.(2018)Peters, Neumann, Iyyer, Gardner, Clark, Lee, and
  Zettlemoyer}]{peters2018}
Matthew~E Peters, Mark Neumann, Mohit Iyyer, Matt Gardner, Christopher Clark,
  Kenton Lee, and Luke Zettlemoyer. 2018.
\newblock \href{https://arxiv.org/pdf/1802.05365.pdf}{Deep contextualized word
  representations}.
\newblock In {\em Proceedings of NAACL-HLT\/}.
\newblock
  \href{https://arxiv.org/pdf/1802.05365.pdf}{https://arxiv.org/pdf/1802.05365.pdf}.

\bibitem[{Pradhan et~al.(2013)Pradhan, Moschitti, Xue, Ng, Bj{\"o}rkelund,
  Uryupina, Zhang, and Zhong}]{pradhan2013}
Sameer Pradhan, Alessandro Moschitti, Nianwen Xue, Hwee~Tou Ng, Anders
  Bj{\"o}rkelund, Olga Uryupina, Yuchen Zhang, and Zhi Zhong. 2013.
\newblock \href{http://www.aclweb.org/anthology/W13-3516}{Towards robust
  linguistic analysis using ontonotes}.
\newblock In {\em Proceedings of the Seventeenth Conference on Computational
  Natural Language Learning\/}. Association for Computational Linguistics,
  pages 143--152.
\newblock
  \href{http://www.aclweb.org/anthology/W13-3516}{http://www.aclweb.org/anthology/W13-3516}.

\bibitem[{Punyakanok et~al.(2005)Punyakanok, Roth, and Yih}]{punyakanok2005}
Vasin Punyakanok, Dan Roth, and Wen-tau Yih. 2005.
\newblock \href{http://cogcomp.org/papers/PunyakanokRoYi05.pdf}{The necessity
  of syntactic parsing for semantic role labeling}.
\newblock In {\em International Joint Conference on Artificial Intelligence
  (IJCAI)\/}. volume~5, pages 1117--1123.
\newblock
  \href{http://cogcomp.org/papers/PunyakanokRoYi05.pdf}{http://cogcomp.org/papers/PunyakanokRoYi05.pdf}.

\bibitem[{Shi et~al.(2016)Shi, Padhi, and Knight}]{shi2016}
Xing Shi, Inkit Padhi, and Kevin Knight. 2016.
\newblock \href{https://doi.org/10.18653/v1/D16-1159}{Does string-based neural
  {MT} learn source syntax?}
\newblock In {\em Proceedings of the 2016 Conference on Empirical Methods in
  Natural Language Processing\/}. Association for Computational Linguistics,
  pages 1526--1534.
\newblock
  \href{https://doi.org/10.18653/v1/D16-1159}{https://doi.org/10.18653/v1/D16-1159}.

\bibitem[{Silveira et~al.(2014)Silveira, Dozat, de~Marneffe, Bowman, Connor,
  Bauer, and Manning}]{silveira14}
Natalia Silveira, Timothy Dozat, Marie-Catherine de~Marneffe, Samuel Bowman,
  Miriam Connor, John Bauer, and Christopher~D. Manning. 2014.
\newblock \href{https://nlp.stanford.edu/pubs/Gold\_LREC14.pdf}{A gold standard
  dependency corpus for {E}nglish}.
\newblock In {\em Proceedings of the Ninth International Conference on Language
  Resources and Evaluation (LREC-2014)\/}.
\newblock
  \href{https://nlp.stanford.edu/pubs/Gold\_LREC14.pdf}{https://nlp.stanford.edu/pubs/Gold\_LREC14.pdf}.

\end{thebibliography}
\bibliographystyle{acl_natbib}

\newpage
\appendix
\onecolumn

\section{Supplementary Materials for Deep RNNs Encode Soft Hierarchical Syntax}
\label{sec:supplemental}
A full list of the accuracy results from the syntax prediction experiments can be found in Table \ref{sup-results}.

\begin{table}[h]
	\centering
	\begin{tabular}{|c|c|c|c|c|c|c|c|}
		\hline
		\multicolumn{3}{|c|}{} & \multicolumn{5}{|c|}{Layer} \\
		\hline
    	Source model & Prediction task & MFT & 0 & 1 & 2 & 3 & 4 \\
    	\hline
    	DP & POS & 0.8801 & \textbf{0.9990} & 0.9964 & 0.9853 & 0.9434 & 0.8962 \\ 
    	& Parent & 0.8190 & 0.8681 & 0.9177 & 0.9347 & \textbf{0.9384} & 0.9349 \\
    	& Grandparent & 0.5422 & 0.5721 & 0.7538 & 0.7920 & 0.8094 & \textbf{0.8253} \\
    	& Great-Grandparent & 0.4511 & 0.4648 & 0.5909 & 0.6659 & 0.6826 & \textbf{0.7183} \\
    	\hline
    	SRL & POS & 0.8801 & 0.8732 & \textbf{0.9063} & 0.8691 & 0.7833 & 0.5392 \\ 
    	& Parent & 0.8190 & 0.7983 & 0.8727 & \textbf{0.8892} & 0.8835 & 0.7870 \\
    	& Grandparent & 0.5422 & 0.5041 & 0.6812 & 0.7394 & \textbf{0.7549} & 0.6325 \\
    	& Great-Grandparent & 0.4511 & 0.4412 & 0.5415 & 0.6159 & \textbf{0.6449} & 0.5493 \\
    	\hline
    	MT & POS & 0.8801 & 0.8618 & \textbf{0.9274} & 0.9198 & 0.9191 & 0.9195 \\ 
    	& Parent & 0.8190 & 0.8019 & 0.8975 & 0.9040 & \textbf{0.9088} & 0.9083 \\
    	& Grandparent & 0.5422 & 0.5368 & 0.7072 & 0.7311 & 0.7572 & \textbf{0.7776} \\
    	& Great-Grandparent & 0.4511 & 0.4361 & 0.5631 & 0.5909 & 0.6303 & \textbf{0.6752} \\
    	\hline
    	LM & POS & 0.8801 & 0.8608 & 0.9093 & \textbf{0.9359} & 0.9304 & 0.9165 \\ 
    	& Parent & 0.8190 & 0.8126 & 0.8724 & \textbf{0.9232} & 0.9137 & 0.9000 \\
    	& Grandparent & 0.5422 & 0.5237 & 0.6251 & \textbf{0.7862} & 0.7606 & 0.7189 \\
    	& Great-Grandparent & 0.4511 & 0.4249 & 0.4702 & \textbf{0.6423} & 0.6302 & 0.5971 \\

    	\hline
	\end{tabular}
	\caption{Table of accuracy results for the syntax feature prediction experiments with best performing layer in each source model/ prediction task pair in bold. ``DP'' refers to the dependency parsing model.}
    \label{sup-results}
\end{table}

\end{document}